# A Comparative Analysis of Deep Reinforcement Learning-enabled Freeway Decision-making for Automated Vehicles

Teng Liu, Yuyou Yang, Wenxuan Xiao, Xiaolin Tang, Mingzhu Yin*

*Abstract*—**Deep reinforcement learning (DRL) has emerged as a pervasive and potent methodology for addressing artificial intelligence challenges. Due to its substantial potential for autonomous self-learning and self-improvement, DRL finds broad applications across various research domains. This article undertakes a comprehensive comparison of several DRL approaches concerning the decision-making challenges encountered by autonomous vehicles on freeways. These techniques encompass common deep Q-learning (DQL), double deep Q-learning (DDQL), dueling deep Q-learning, and prioritized replay deep Q-learning. Initially, the reinforcement learning (RL) framework is introduced, followed by a mathematical establishment of the implementations of the aforementioned DRL methods. Subsequently, a freeway driving scenario for automated vehicles is constructed, wherein the decision-making problem is reformulated as a control optimization challenge. Finally, a series of simulation experiments are conducted to assess the control performance of these DRL-enabled decision-making strategies. This culminates in a comparative analysis, which seeks to elucidate the connection between autonomous driving outcomes and the learning characteristics inherent to these DRL techniques.**

## NOMENCLATURE

| | |
|---|---|
| HSM | Hierarchical State Machine |
| CTP | Critical Turning Point |
| POMDP | Partially Observable Markov Decision Process |
| DRL | Deep Reinforcement Learning |
| DL | Deep Learning |
| RL | Reinforcement Learning |
| DQL | deep Q learning |
| PPO | Proximal Policy Optimization |
| A3C | Asynchronous Advantage Actor-Critic |
| IDM | Intelligent Driver Model |
| MOBIL | Minimizing Overall Braking Induced by Lane Change |
| DDQL | double DQL |
| MDP | Markov decision process |

This work was supported by National Key R&D Programmes (NKPs) of China (No. 2022YFC3601800), the hospital-level scientific research project of Chongqing University Three Gorges Hospital (No. 2023YJKYXM-001), and the major project of science and technology research program of Chongqing Education Commission of China (No. KJZD-K202300105). (Corresponding authors: M. Yin)

Y. Yang, W. Xiao, and X. Tang are with College of Mechanical and Vehicle Engineering, Chongqing University, Chongqing, 400044, China. (email: yangyy@stu.cqu.edu.cn, wen-xuanxiao@cqu.edu.cn, tangxiaolin6@126.com)

Teng Liu is with Clinical Research Center and Medical Pathology Center, Chongqing University Three Gorges Hospital, Chongqing University, Wanzhou 404000, Chongqing, China, and also with College of Mechanical and Vehicle Engineering, Chongqing University, Chongqing 400044, China. (email: tengliu17@cqu.edu.cn)

M. Yin is with Clinical Research Center, Clinical Pathology Center, and Cancer Early Detection and Treatment Center, Chongqing University Three Gorges Hospital, Chongqing University, Wanzhou, Chongqing, China. (email: yinmingzhu2008@126.com)

## I. INTRODUCTION

Autonomous driving has recently garnered significant attention due to its substantial potential to enhance both driving safety and efficiency [1]. Remarkable progress has been achieved in the development of an autonomous driving system's four pivotal components, namely, perception, decision-making, planning, and control [2].

The decision-making module serves as the central intelligence hub within the autonomous driving system [3]. Its primary responsibility is to formulate appropriate motion behaviors tailored to specific missions within a dynamic and uncertain environment. Numerous research endeavors have been dedicated to enhancing the decision-making module's capabilities in the realm of autonomous driving. As an illustration, Q. Do et al. introduced a grid map for categorizing the driving environment. To select suitable behaviors for each category, the authors constructed a dx/dv graph based on real-world data [4]. In the context of Ref. [5], a hierarchical state machine is employed to address the behavior decision-making challenge associated with lane-changing within predefined environments. Additionally, Shu, K. et al. proposed a hierarchical decision-making strategy for left turns at intersections. The high-level layer focuses on path generation based on critical turning points, while the low-level layer transforms the left-turn planning problem into a partially observable Markov decision process (POMDP) problem using the concept of Critical Turning Point (CTP) [6]. Furthermore, in [7], the authors utilized a mixed observable Markov decision process to model the decision-making process for lane-changing, taking into consideration the motion uncertainties of vehicles.

Deep Reinforcement Learning (DRL) stands out as a promising approach for autonomous driving decision-making. DRL harnesses the robust feature extraction capabilities inherent to Deep Learning (DL) and the decision-making prowess of Reinforcement Learning (RL) in sequential problem-solving contexts. For instance, Mirchevska B et al. introduced a DQL-based approach for behavior decision-making during lane-changing scenarios within a simulated highway environment [8]. To ensure safety, they defined a formal safety verification process. In Ref. [9], a hierarchical RL-based control architecture was proposed, capable of generating safe and smooth trajectories without heavy reliance on extensive labeled driving data. Subsequently, F. Ye et al. devised a Proximal Policy Optimization (PPO)-based DRL strategy for mandatory lane-changing, demonstrating a remarkable 95% action accuracy even in congested traffic conditions [10]. In Ref. [11], X. Feng et al. constructed a lane-changing strategy using DRL, encompassing both DQL and Asynchronous Advantage Actor-Critic (A3C), showcasing excellent generalization capabilities. To enhance its performance, a novel approach for combining DRL agents was introduced. As of our current knowledge, there exists no comprehensive literature comparing the performance of various DRL methods in the context of freeway decision-making.

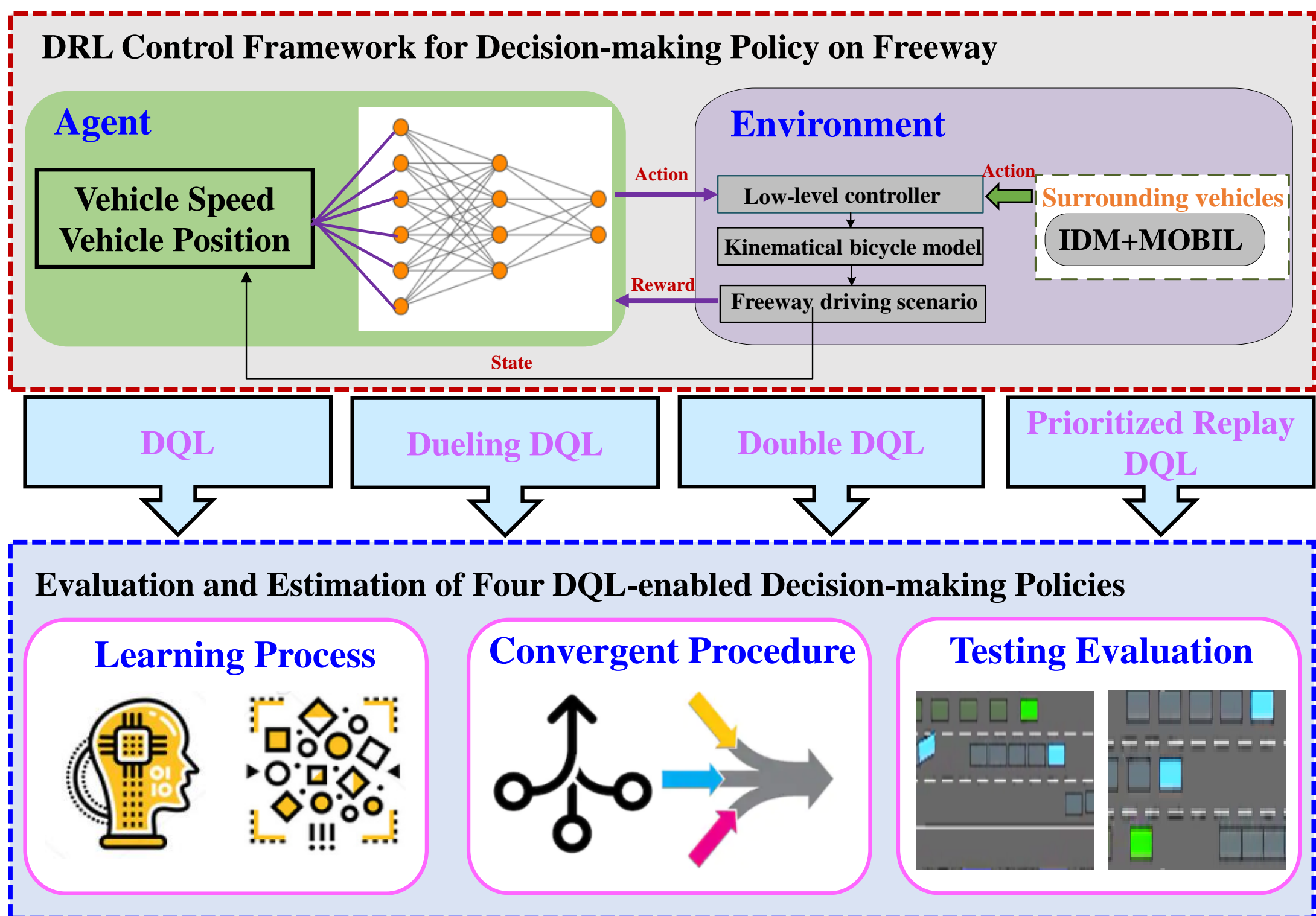

Fig. 1. Decision-making control framework based on four DQL algorithms for autonomous vehicles on freeway.

In this study, we undertake a comprehensive comparison of various Deep Reinforcement Learning (DRL) approaches concerning the decision-making challenges faced by autonomous vehicles on freeways. These approaches encompass Deep Q Learning (DQL), Double Deep Q Learning (DDQL), Dueling Deep Q Learning, and Prioritized Replay Deep Q Learning, as visually represented in Figure 1. Initially, we introduce the Reinforcement Learning (RL) framework and elucidate the DQL methods mentioned earlier. Subsequently, we formulate the decision-making quandary encountered by autonomous vehicles in the context of freeway driving, wherein the surrounding vehicles adhere to the Intelligent Driver Model (IDM) and the Minimizing Overall Braking Induced by Lane Change (MOBIL) model. Finally, we conduct simulation experiments to facilitate a comprehensive discussion and analysis of the control performance exhibited by these compared DRL-enabled decision-making strategies.

This paper enumerates its potential contributions as follows: 1) A meticulous and comprehensive comparison is undertaken, focusing on epidemic Deep Reinforcement Learning (DRL) techniques applied to decision-making challenges in freeway scenarios. 2) The paper conducts a thorough theoretical and mathematical analysis to formulate the Deep Q Learning (DQL), Double Deep Q Learning (DDQL), Dueling Deep Q Learning, and Prioritized Replay Deep Q Learning (DQL) algorithms. 3) It provides a comprehensive discussion and evaluation, elucidating both the advantages and disadvantages inherent in various DRL-based decision-making policies. This work aspires to serve as a guiding framework for the application of DRL approaches to address decision-making challenges in the context of freeway travel for autonomous vehicles.

The structure of this paper is delineated as follows: Multiple Deep Q Learning (DQL) methods are presented in Section 2. Section 3 formulates the decision-making challenges encountered on freeways. The experimental results are compared and analyzed in Section 4. Finally, Section 5 provides the concluding remarks of the paper.

## II. Introduction of Multiple DQL Methods

This section serves to introduce various Deep Reinforcement Learning (DRL) methods. Initially, it provides an explanation of the Markov decision process and elucidates the interaction dynamics between the agent and the environment within the realm of Reinforcement Learning (RL). Subsequently, Common Deep Q Learning (DQL) is introduced, a method that amalgamates a neural network with a Q-learning algorithm. Lastly, this section delves into the presentation of three en-hanced iterations of DQL, namely, Double Deep Q Learning, Dueling Deep Q Learning, and Prioritized Replay Deep Q Learning, each of which is introduced individually.

### *A. Preliminaries of RL*

The challenge of addressing Reinforcement Learning (RL) can be comprehended as the endeavor to maximize the cumulative reward attained by the agent during its interactions with the environment [12-15]. In cases where the environment is entirely observable, individuals can formulate a Markov decision process (MDP) to encapsulate the entirety of the reinforcement learning problem. However, situations may arise in which the environment is not entirely observable, leading individuals to construct an approximate, fully observable environment description based on historical observation data. Consequently, it can be asserted that nearly all RL problems can be reformulated as MDPs [16-18].

In the context of the highway decision-making problem, the agent corresponds to the ego vehicle, while the environment encompasses the surrounding vehicles, including the prevailing driving conditions. The essence of the Markov property lies in the notion that each state within the process exhibits this property; that is to say, the state variable in the subsequent moment solely relies on the current state and bears no connection to the state at prior time instances [19]. Markov Decision Processes (MDPs) form the basis of modeling such scenarios, embracing the Markov property and taking into account the feedback from state transitions and the selection of individual behaviors. MDPs are capable of representing Reinforcement Learning (RL) problems in the form of tuples (S, A, P, R, γ), where S and A denote the sets of states and actions, respectively. P constitutes the crux of RL, signifying the state transition model grounded in behavior, while R represents the reward model contingent upon state and behavior. To account for the diminished importance of future rewards relative to present rewards, the discount rate, denoted as γ, is introduced. This rate quantifies the significance assigned to future rewards in the decision-making process.

The cumulative return $R$ reflects the sum of future rewards, and can be calculated as follows:

$$R=\sum_{t=0}^{K}\gamma^{t}r_{t} \tag{1}$$

where $t$ is the discrete-time steps to make control decisions in an episode, and $r_t$ indicates the relevant reward. $K$ is the time-step of the end of the environment, which could be ∞.

Value functions could be used to learn policies. These functions can calculate the expectation of $R$ under a given strategy. Two value functions are defined based on the state s and state-action pair ($s$, $a$) as:

$$V^{\pi}(s)=E_{\pi}[\sum_{t=0}^{K}\gamma^{t}r_{t}\mid s_{t=0}=s] \tag{2}$$

$$Q^{\pi}(s,a)=E_{\pi}[\sum_{t=0}^{K}\gamma^{t}r_{t}\mid s_{t=0}=s,a_{k=0}=a] \tag{3}$$

where $\pi$ is the policy, $V$ is the state value function, and $Q$ is the state-action value function. The state-action function can also be written in another recursive form:

$$Q^{\pi}(s,a)=E_{\pi}[r_{t}+\gamma\max_{a'}Q^{\pi}(s',a')] \tag{4}$$

where $s'$ and $a'$ are the state and action at the next time step.

Finally, the optimal control policy is determined by selecting the action that yields the highest expected return value, as computed through the state-action value function.

$$\pi(s)=\arg\max_{a}Q(s,a) \tag{5}$$

### *B. Common DQL*

DQL integrates Q-learning with deep learning, employing a neural network to estimate Q-values. In conventional Q-learning, the recursive form of the state-action function is expressed as follows:

$$Q(s,a)\leftarrow Q(s,a)+\alpha[r+\gamma\max_{a'}Q(s',a')-Q(s,a)] \tag{6}$$

where $\alpha$ is the learning rate.

When confronted with numerous state and action variables, Common Q-learning requires a substantial amount of time for the computation of the mutable Q-table. DQL adeptly addresses this issue by employing a neural network to represent the Q-table as Q(s, a; θ), where θ represents the parameters of the neural network. The central objective of the DQL method revolves around the precise training of these parameters.

Q-learning employs an ε-greedy policy for action generation. It's worth noting that the strategy for generating behavior and the strategy for evaluation can be distinct, a concept referred to as the Off-Policy method. Deep Q Learning (DQL) also falls into the category of Off-Policy algorithms. However, a notable distinction exists between Q-learning and DQL: Q-learning relies on a single neural network to calculate target and predicted values, whereas DQL utilizes two networks—namely, the prediction network and the target network. Both networks share the same architectural structure, but the prediction network undergoes updates in each iteration, whereas the target network periodically copies the parameters from the prediction network after a specified number of time steps.

To capture the disparity between the approximate Q-table and the true Q-table in Deep Q Learning (DQL), a loss function is introduced as follows:

$$L(\theta)=E[\sum\nolimits_{t=1}^{N}(y_{t}-Q(s,a;\theta))^{2}] \tag{7}$$

where

$$y_{t}=r_{t}+\gamma\max_{a'}Q(s',a';\theta') \tag{8}$$

The $\theta$ and $\theta'$ represent the parameters of the prediction network and target network, respectively.

The neural network uses a gradient descent method to update iteratively in DQL, and the gradient formula is as follows:

$$\nabla_{\theta}L(\theta)=E[(y_{i}-Q(s,a;\theta))\nabla_{\theta}Q(s,a;\theta)] \tag{9}$$

In addition, DQL incorporates the experience replay method [20]. Given the substantial correlation among samples, direct learning from continuous samples proves to be inefficient. To mitigate this, the experience replay method is employed to introduce randomness into the samples, thereby disrupting correlations and enhancing learning efficiency. The Common DQL algorithm is delineated in Table I, where the function Φ denotes the utilization of a fixed representation length of historical data as the input to the neural network.

TABLE I
IMPLEMENTATION CODE OF COMMON DQL ALGORITHM

| **Common DQL Algorithm** |
|---|
| **1.** Initialize replay memory $D$ to capacity $N$ |
| **2.** Initialize action-value function $Q$ with random weights |
| **3.** **For** episode = 1, 2, …, $M$ **do** |
| **4.** Initialize state $s_1$ and preprocessed sequenced $\Phi_1$ |
| **5.** **For** $t$ = 1, 2, …, $T$ **do** |
| **6.** With probability $\varepsilon$ select a random action $a_t$<br>otherwise select $a_t = \max_a Q(\Phi(s_t), a; \theta)$ |
| **7.** Execute action at in emulator and observe reward $r_t$ and image $x_{t+1}$ |
| **8.** Set $s_{t+1} = s_t, a_t, x_{t+1}$ and preprocess $\Phi_{t+1} = \Phi(s_{t+1})$ |
| **9.** Store transition $(\Phi_t, a_t, r_t, \Phi_{t+1})$ in $D$ |
| **10.** Sample random minibatch of transitions $(\Phi_j, a_j, r_j, \Phi_{j+1})$ from $D$ |
| **11.** Set $y_j = r_j$ for terminal $\Phi_{j+1}$<br>otherwise set $y_j = r_j + \gamma \max_{a'} Q(\Phi_{j+1}, a'; \theta)$ |
| **12.** Perform a gradient descent step on $(y_j - Q(\Phi_j, a_j; \theta))^2$ according to equation 9 |
| **13.** **end for** |
| **14.** **end for** |

## C. Double DQL

Both Common Q-learning and DQL employ the max operator for value function assessment and action selection. However, this optimistic selection approach can result in value overestimation [21]. To counter this issue, a strategy involves employing two distinct value functions to decouple the action selection from the evaluation process. Consequently, in the case of Double Deep Q Learning (DDQL), the max operation is disassembled into action selection and action evaluation.

DQL itself encompasses two neural networks, namely the prediction network and the target network, obviating the need for additional networks in DDQL. Beyond the calculation method of the target value, the algorithmic flow of the Double DQL and Common DQL remains identical. The target value ($y_t$) in Double DQL is expressed as follows:

$$y_t^{Double\,DQL} = r_t + \gamma Q(s', \arg\max_{a'} Q(s', a'; \theta), \theta') \tag{10}$$

Therefore, it becomes possible to assess the greedy strategy based on the prediction network and employ the target network to estimate its value. This DDQL algorithm effectively mitigates the issue of value overestimation. The procedural implementation of Double DQL is illustrated in Table II.

TABLE II
IMPLEMENTATION CODE OF DOUBLE DQL ALGORITHM

| **Double DQL Algorithm** |
|---|
| **1.** Input: target network replacement frequence $N'$ |
| **2.** Initialize replay memory $D$ to capacity $N$ |
| **3.** Initialize action-value function $Q$ with random weights |
| **4.** **For** episode = 1, 2, …, $M$ **do** |
| **5.** Initialize state $s_1$ and preprocessed sequenced $\Phi_1$ |
| **6.** **For** $t$ = 1, 2, …, $T$ **do** |
| **7.** With probability $\varepsilon$ select a random action $a_t$<br>otherwise select $a_t = \max_a Q(\Phi(s_t), a; \theta)$ |
| **8.** Execute action at in emulator and observe reward $r_t$ and image $x_{t+1}$ |
| **9.** Set $s_{t+1} = s_t, a_t, x_{t+1}$ and preprocess $\Phi_{t+1} = \Phi(s_{t+1})$ |
| **10.** Store transition $(\Phi_t, a_t, r_t, \Phi_{t+1})$ in $D$ |
| **11.** Sample random minibatch of transitions $(\Phi_j, a_j, r_j, \Phi_{j+1})$ from $D$ |
| **12.** Set $y_j = r_j$ for terminal $\Phi_{j+1},$<br>otherwise set $y_j = r_j + \gamma Q(\Phi_{j+1}, \mathrm{argmax}_{a'} Q(\Phi_{j+1}, a'; \theta); \theta')$ |
| **13.** Perform a gradient descent step on $(y_j - Q(\Phi_j, a_j; \theta))^2$ according to equation 9 |
| **14.** Replace target parameters $\theta$ *by* $\theta'$ every $N'$ steps |
| **15.** **end for** |
| **16.** **end for** |

## D. Dueling DQL

In certain RL scenarios, there exist states in which the choice of action, regardless of its nature, will have minimal impact on the subsequent state. For instance, when the highway ahead of the agent is devoid of other vehicles, the action selection does not significantly alter the driving state. To account for this, Dueling Deep Q Learning (DQL) decomposes the state-action function into two distinct components: the state-value function, denoted as V(s), and the advantage function, denoted as A(s, a). This decomposition proves more adept at capturing the nuances of the reinforcement learning process. The expressions are as follows:

$$Q^{\pi}(s,a) = A^{\pi}(s,a) + V^{\pi}(s) \tag{11}$$

Dueling DQL diverges from Common DQL by adopting a different network architecture. Rather than appending fully connected layers directly after the convolutional layers, Dueling DQL employs a stream of two fully connected layers. This distinctive network structure enables the separate estimation of the value and advantage functions. Ultimately, the single output of the Q function is derived by amalgamating the outcomes of these two streams. The calculation of the state-action function, featuring two parameters in the network, is articulated as follows:

$$Q^{\pi}(s,a;\theta,\beta_1,\beta_2) = V^{\pi}(s;\theta,\beta_1) + A^{\pi}(s,a;\theta,\beta_2) \tag{12}$$

where $\theta$ is the parameters of the convolutional layers, while $\beta_1$ and $\beta_2$ are the parameters of the two streams of fully-connected layers.

Equation (12) presents an issue of unidentifiability, indicating that we are unable to uniquely retrieve V and A from the provided Q. To address this identifiability challenge, a solution entails constraining the advantage function estimator to possess zero advantage in the selected actions [22]. Consequently, (12) can be redefined as follows:

$$Q^{\pi}(s,a;\theta,\beta_1,\beta_2)=V^{\pi}(s;\theta,\beta_1)+ \\ (A^{\pi}(s,a;\theta,\beta_2)-\max_{a'} A^{\pi}(s,a';\theta,\beta_2)) \quad (13)$$

$$a^{*}=\arg\max_{a'} Q(s,a;\theta,\beta_1,\beta_2)=\arg\max_{a'} A(s,a';\theta,\beta_2) \quad (14)$$

By equation (13) and equation (14), we can obtain the following equation:

$$Q^{\pi}(s,a^{*};\theta,\beta_1,\beta_2)=V^{\pi}(s;\theta,\beta_1) \quad (15)$$

Therefore, the reformulated equation (13) possesses identifiability. Moreover, due to the fact that the architecture of Dueling Deep Q Learning (DQL) shares the same input-output interface as Common DQL, training Dueling DQL through Q network-based learning algorithms is a straightforward process. The pseudocode for Dueling DQL is provided in Table III, wherein the advantage network is capable of estimating the value of each selected control action. This attribute enhances the quality of the resulting control policy.

TABLE III
IMPLEMENTATION CODE OF DUELING DQL ALGORITHM

| **Dueling DQL Algorithm** |
|---|
| **1.** Initialize replay memory $D$ to capacity $N$ |
| **2.** Initialize action-value function $Q$ with random weights |
| **3.** **For** episode = 1, 2, …, $M$ **do** |
| **4.** Initialize state $s_1$ and preprocessed sequenced $\Phi_1$ |
| **5.** **For** $t$ = 1, 2, …, $T$ **do** |
| **6.** With probability $\varepsilon$ select a random action $a_t$ otherwise select $a_t = \max_a Q(\Phi(s_t), a; \theta)$ |
| **7.** Execute action at in emulator and observe reward $r_t$ and image $x_{t+1}$ |
| **8.** Set $s_{t+1} = s_t, a_t, x_{t+1}$ and preprocess $\Phi_{t+1} = \Phi(s_{t+1})$ |
| **9.** Store transition $(\Phi_t, a_t, r_t, \Phi_{t+1})$ in $D$ |
| **10.** Sample random minibatch of transitions $(\Phi_j, a_j, r_j, \Phi_{j+1})$ from $D$ |
| **11.** Calculate two streams of evaluated deep networks including $V(\Phi(s_t); \theta, \beta_1)$ and $A(\Phi(s_t), a; \theta, \beta_2)$ and combine them as $Q(\Phi(s_t), a; \theta, \beta_1, \beta_2)$ using equation (13) |
| **12.** Set $y_j = r_j$ for terminal $\Phi_{j+1}$ otherwise set $y_j = r_j + \gamma \max_{a'} Q(\Phi_{j+1}, a'; \theta)$ |
| **13.** Perform a gradient descent step on $(y_j - Q(\Phi_j, a_j; \theta))^2$ according to equation 9 |
| **14.** **end for** |
| **15.** **end for** |

*E. Prioritized Replay DQL*

In the case of Common Deep Q Learning (DQL), the sampling of learning is uniform and random. However, it is worth noting that in certain Reinforcement Learning (RL) scenarios, this approach may yield suboptimal learning efficiency. To address this concern, Prioritized Replay DQL incorporates the concept of temporal-difference (TD) error δ. TD error serves as a metric that quantifies the disparity between the value in the current state and the estimate for the next step. This metric offers an approximation of the extent to which the agent can gain knowledge from a transition within its present state [23]. The calculation of TD error δ is expressed as follows:

$$\delta=| r+\gamma \max_{a'} Q(s',a')-Q(s,a) | \quad (16)$$

Then the sample priorities are arranged in ascending order based on the value of δ. The greater the priority assigned to a sample, the higher the likelihood of it being selected for the learning process. The probability of selecting transition i through sampling is defined as follows:

$$P(i)=\frac{p_i^{\psi}}{\sum_k p_i^{\psi}} \quad (17)$$

where $p_i > 0$ is the priority of transition $i$. The exponent $\psi$ determines how much prioritization is used. In proportional prioritization, the expression of $p_i$ is as follows：

$$p_i=| \delta_i |+\varepsilon \quad (18)$$

where $\varepsilon$ is a small positive constant. $\varepsilon$ is added in the equation (18) to prevent situations that the probability of sampling transition is zero when the TD error is zero.

TABLE IV
IMPLEMENTATION CODE OF PRIORITIZED REPLAY DQL ALGORITHM

| **Prioritized Replay DQL Algorithm** |
|---|
| **1.** Input: minibatch $k$, step-size $\eta$, replay period $K$ and size $N$, exponents $\psi$ and $\lambda$, budget $T$. |
| **2.** Initialize replay memory $H$, $\Delta = 0$, $p_1 = 1$ |
| **3.** Observe $S_0$ and choose $A_0 \sim \pi_\theta(S_0)$ |
| **4.** **For** $t$ = 1,2, …, $T$ **do** |
| **5.** Observe $S_t, R_t, \gamma_t$ |
| **6.** Store transition $(S_{t-1}, A_{t-1}, R_t, \gamma_t, S_t)$ in H with maximal priority $p_t = \max_{i<t} p_i$ |
| **7.** **if** $t \equiv 0$ mod K then |
| **8.** **For** $j$ = 1,2, …, $k$ **do** |
| **9.** Sample transition $j \sim P(j) = p_j^{\psi} / \Sigma_i p_i^{\psi}$ |
| **10.** Compute importance-sampling weight $\omega_j = (N \cdot P(j))^{-\lambda} / \max_i \omega_i$ |
| **11.** Compute TD-error $\delta_j$ according to equation 16 |
| **12.** Update transition priority $p_j \leftarrow \|\delta_j\|$ |
| **13.** Accumulate weight-change $\Delta \leftarrow \Delta + \omega_j \cdot \delta_j \cdot \nabla_\theta Q(S_{j-1}, A_{j-1})$ |
| **14.** **end for** |
| **15.** Update weights $\theta \leftarrow \theta + \eta \cdot \Delta$, reset $\Delta = 0$ |
| **16.** **end if** |
| **17.** Choose action $A_t \sim \pi_\theta(S_t)$ |
| **18.** **end for** |

Stochastic prioritization serves as a solution to address the potential loss of diversity that might arise when samples are selected based on priority. However, this approach also introduces a degree of bias. To mitigate this bias, importance-sampling (IS) weights can be employed:

$$\omega_i=(\frac{1}{N}\cdot\frac{1}{P(i)})^{\lambda} \quad (19)$$

where N is the size of memory and it is the exponent. If $\lambda$=1, the non-uniform probabilities can be fully compensated by using $\omega_i \delta_i$ instead of $\delta_i$. Finally, the Prioritized Replay Deep Q Learning (DQL) algorithm is obtained, which enhances learning speed through the optimization of the random sampling procedure. The implemented code for Prioritized Replay DQL is delineated in Table IV.

## III. Decision-making Problem on Freeway

In this section, we outline the construction of the decision-making problem within the freeway context. Initially, we establish the freeway driving scenario and elucidate the training objectives. Subsequently, we introduce both the high-level controller, which operates in conjunction with the IDM and the MOBIL, to govern the behavior of surrounding vehicles, and the low-level controller, which oversees all vehicles and generates motion commands. Additionally, we employ a kinematic bicycle model to comprehensively capture the motion characteristics of the vehicles involved in the scenario.

### A. *Freeway Driving Scenario*

The decision-making module plays a pivotal role in shaping the driving performance of autonomous vehicles [24]. Within this module, autonomous vehicles are tasked with the selection of suitable driving behaviors and the formulation of safe and efficient trajectories to tracking.

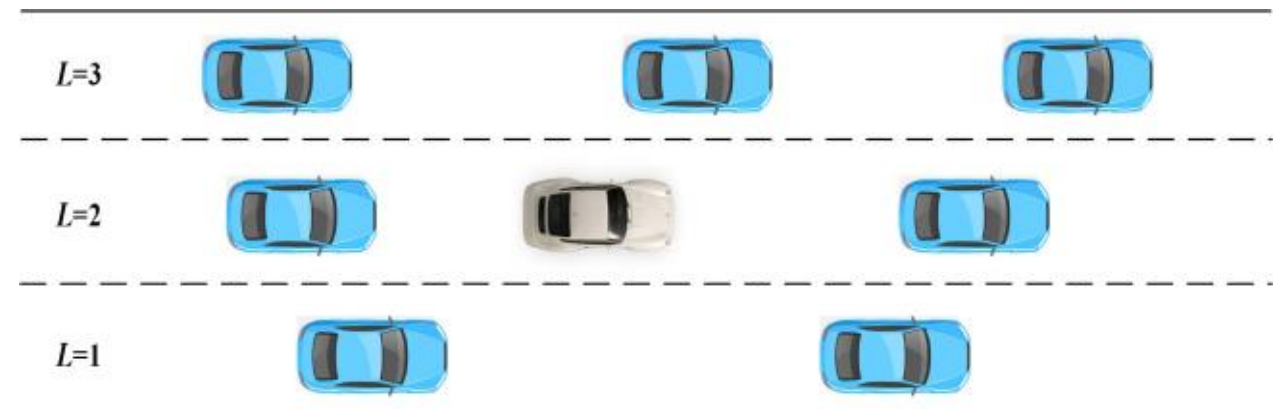

Fig. 2. The freeway driving scenario.

This study selects a segment of the freeway as the driving scenario, wherein common driving behaviors primarily encompass car-following, lane-changing, and overtaking maneuvers. The visual representation of this scenario is provided in Fig. 2, with the white car symbolizing the ego vehicle and the blue cars representing the surrounding vehicles. In this particular setting, the number of lanes and surrounding vehicles is denoted as K and N, respectively.

The objective for the ego vehicle is to achieve and sustain the highest possible speed while avoiding collisions with other vehicles. Furthermore, the ego vehicle is encouraged to occupy the rightmost lane when not engaged in overtaking maneuvers. Throughout the training process, the ego vehicle endeavors to acquire an effective decision-making strategy that aligns with these driving objectives.

Within each episode, both the initial velocities of the ego vehicle and surrounding vehicles are randomly selected from the ranges [23, 25] m/s and [20, 23] m/s, respectively. The maximum attainable speed is capped at 40 m/s, with all vehicles sharing a common length and width of 5m and 2m, respectively. Each episode has a duration of 100 seconds, but it concludes immediately in the event of a collision.

### B. *High-level Controller of Surrounding Vehicles*

In this study, the anticipated speed and target lane for the ego vehicle at each time step are determined through DRL. Meanwhile, the behaviors of the surrounding vehicles are regulated in accordance with the IDM and the MOBIL model [25] [26]. Specifically, IDM is employed to compute longitudinal acceleration, ensuring car-following is carried out safely without collisions, while MOBIL assesses the feasibility and value of lane-changing intentions to facilitate lane changes. The default parameter settings for IDM and MOBIL are outlined in Table V.

TABLE V
The Default Configuration of the IDM and MOBIL

| Symbol | Meaning | Values |
|---|---|---|
| $a_{\max}$ | Maximum acceleration | 6 m/s$^2$ |
| $\delta$ | Acceleration argument | 4 |
| $d_0$ | Minimum relative distance | 10 m |
| $T$ | Safe time gap | 1.5 s |
| $b$ | Comfortable deceleration rate | 5 m/s$^2$ |
| $b_{safe}$ | Safe deceleration limit | 2 m/s$^2$ |
| $p$ | Politeness factor | 0.001 |
| $a_{th}$ | Acceleration threshold | 0.2 m/s$^2$ |

IDM is commonly employed to regulate the longitudinal acceleration in the context of adaptive cruise control for autonomous vehicles. This acceleration, denoted as a, is computed based on the relative speed (Δv) and the relative distance (Δd) to the vehicle in the front, as expressed below:

$$a = a_{\max}[1-(\frac{v}{v_{ex}})^{\delta}-(\frac{d_{ex}}{\Delta d})^2] \tag{20}$$

where $v$ is the longitude speed of the vehicle at this moment. $a_{max}$ is the maximum acceleration, and $\delta$ is the constant acceleration parameter. $v_{ex}$ is the expected speed derived by $a_{max}$ and $d_{ex}$. $d_{ex}$, the expected distance between the vehicle and its front vehicle, is calculated as:

$$d_{ex} = d_0 + Tv + \frac{v\Delta v}{2\sqrt{a_{\max} b}} \tag{21}$$

where $d_0$ and $T$ are the minimum relative distance and the time interval, which are predefined to ensure safety. $b$ is the deceleration rate to improve driving comfort.

MOBIL initiates a lane change when both safety and incentive conditions are met for the target lane. These conditions are defined as follows:

$$a_n^{af} \geq -b_{safe} \tag{22}$$

$$a_e^{af} - a_e^{be} + p[(a_n^{af} - a_n^{be}) + (a_o^{af} - a_o^{be})] \geq a_{th} \tag{23}$$

where $a^{be}{}_e$, $a^{be}{}_o$, $a^{be}{}_n$ are the accelerations of the vehicle, its follower at the initial lane, its follower at the target lane before lane changing. $a^{af}{}_e$, $a^{af}{}_o$, $a^{af}{}_n$ are the accelerations of the vehicle, its follower at the initial lane, its follower at the target lane after lane changing. $b_{safe}$ is the deceleration limit and $a_{th}$ is the acceleration threshold. $p$ is named as the politeness coefficient to meet a trade-off between the vehicle and its followers.

To prevent collisions when a vehicle is changing lanes, it may be necessary for the following vehicle in the target lane to decelerate. The safety condition stipulates that this deceleration must not exceed a certain limit to maintain safety. On the other hand, the incentive condition necessitates that the combined benefit of the vehicle's acceleration and that of its followers surpasses a predefined threshold.

### C. Low-level Controller of Vehicles

Given input parameters such as reference speed and target lane, the low-level controller is responsible for converting these inputs into vehicle acceleration and steering angle. The vehicle's acceleration (a) is regulated using a proportional controller, and its calculation is as follows:

$$a = K_p(v_{ex} - v) \tag{24}$$

where $K_p$ is the acceleration control gain, $v_{ex}$ is the expected speed. The controller of steering angle $\delta$ is a proportional-derivative controller:

$$v_{ex,lat} = -K_{p,lat} d_{lat} \tag{25}$$

$$\theta_{ex} = \arcsin\left(\frac{v_{ex,lat}}{v}\right) + \theta_L \tag{26}$$

$$\dot{\theta} = K_{p,\theta}(\theta_{ex} - \theta) \tag{27}$$

$$\delta = \arcsin\left(\frac{1}{2}\frac{l_r}{v}\dot{\theta}\right) \tag{28}$$

where $K_{p,lat}$ and $K_{p,\theta}$ are the position and heading control gains, $v_{ex,lat}$ is the expected lateral speed, $d_{lat}$ is the lateral distance between the vehicle and the center-line of the target lane, $\theta_L$ is the target lane heading, $\theta_{ex}$ is the expected heading, and $\theta$ is the current heading.

### D. Kinematics of Vehicles

In this study, we employ a Kinematic Bicycle Model [27] to model and analyze vehicle motion. Within this proposed bicycle model, the individual right and left wheels of the vehicle are consolidated into a single wheel to represent vehicle movement, as depicted in Fig. 3. Additionally, this model posits that steering is exclusive to the front wheel, and it assumes that there is no wheel slippage.

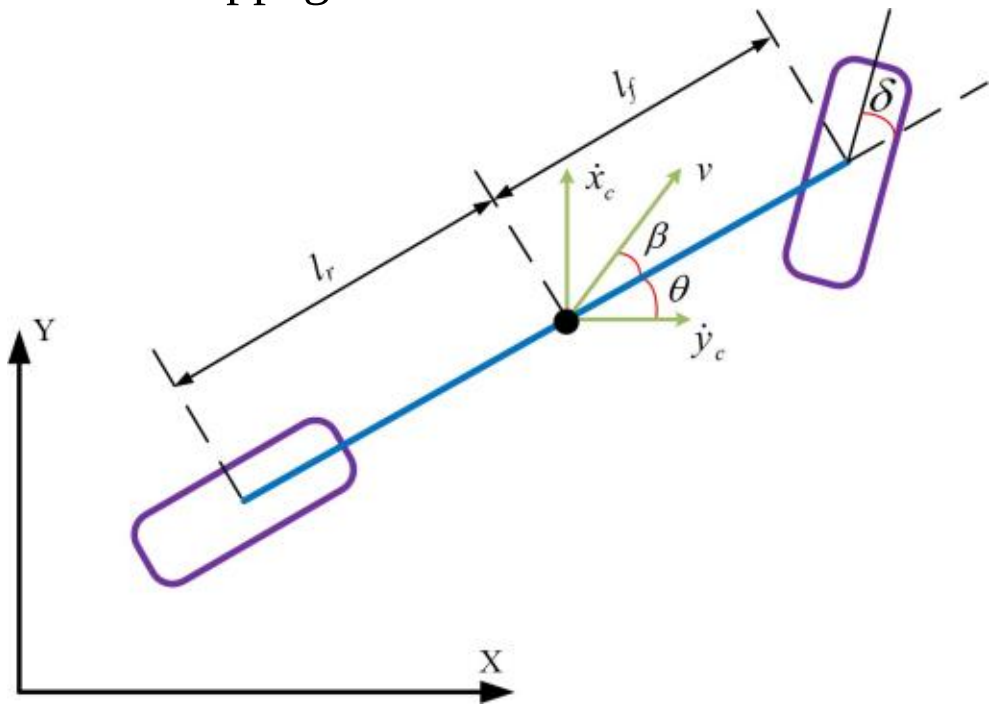

Fig. 3. The kinematic bicycle model.

The model takes two inputs into account: the steering angle of the front wheel (δ) and the acceleration (a). The motion trajectories of the vehicle's center of gravity can be derived based on these inputs, as follows:

$$\dot{x}_c = v\cos(\theta + \beta) \tag{29}$$

$$\dot{y}_c = v\sin(\theta + \beta) \tag{30}$$

$$a = \dot{v} \tag{31}$$

$$\dot{\theta} = \frac{v\sin\beta}{l_r + l_f} \tag{32}$$

$$\beta = \tan^{-1}\left(\frac{l_r \tan\delta}{(l_r + l_f)}\right) \tag{33}$$

where ($x_c$, $y_c$) is the position of the center of gravity, $\theta$ is the heading of the bicycle, $\beta$ is the slip angle at the center of gravity, $l_r$ is the distance between the center of the front wheel and the center of gravity, $l_f$ is the distance between the center of the rear wheel and the center of gravity. The realization of four DQL-based decision-making problem is conducted in Python [28] with the highway environment.

## IV. Experiments and Discussion

This section focuses on the simulation results obtained from four Deep Reinforcement Learning (DRL) methods, with an emphasis on discussion and analysis. Initially, a comparative assessment is made regarding the learning process of these methods, encompassing evaluations of rewards and state variables. Subsequently, the convergence patterns of the DRL algorithms are elucidated. The advantages stemming from the variations within the Deep Q Learning (DQL) technique are expounded upon. Finally, an evaluative experiment is carried out to gauge the applicability of these trained decision-making strategies in a similar test scenario.

### A. Learning Process of DRL Methods

For the sake of convenience, the methods introduced in Section II are referred to as DQL, Double DQL, Dueling DQL, and PR-DQL in the subsequent discussions. The Double DQL algorithm is devised to mitigate the issue of value function overestimation, while Dueling DQL is formulated to emphasize the advantages of control action selection. PR-DQL is adopted to enhance learning efficiency through the prioritization of select sample experiences. Across these four methods, identical parameter settings are applied. The total number of epochs is fixed at 2000, and the duration of the driving scenario is set to 100 (signifying that the maximum attainable reward value is 100). The discount factor (γ) and learning rate (α) are configured as 0.8 and 0.2, respectively. The road comprises three lanes, and there are a total of 15 surrounding vehicles. The reward function is composed of a combination of vehicle speeds and collision conditions.

This subsection elucidates the learning process of the four algorithms within the context of the freeway decision-making problem. In Fig. 4, various types of normalized reward trajectories are presented, encompassing raw data points, original trend lines, smoothed trend lines, and linear fits. These trajectories depict an upward trend in reward values as the number of episodes progresses, signifying that the ego vehicle progressively gains familiarity with the driving environment through a process of trial and error. Upon closer examination, it becomes evident that Dueling DQL consistently exhibits the best performance, consistently surpassing the other three methods in terms of rewards. In the cases of smooth trend lines and linear fits, Dueling DQL consistently outperforms the other three methods. Both Double DQL and PR-DQL demonstrate similar performance, which is superior to that of DQL.

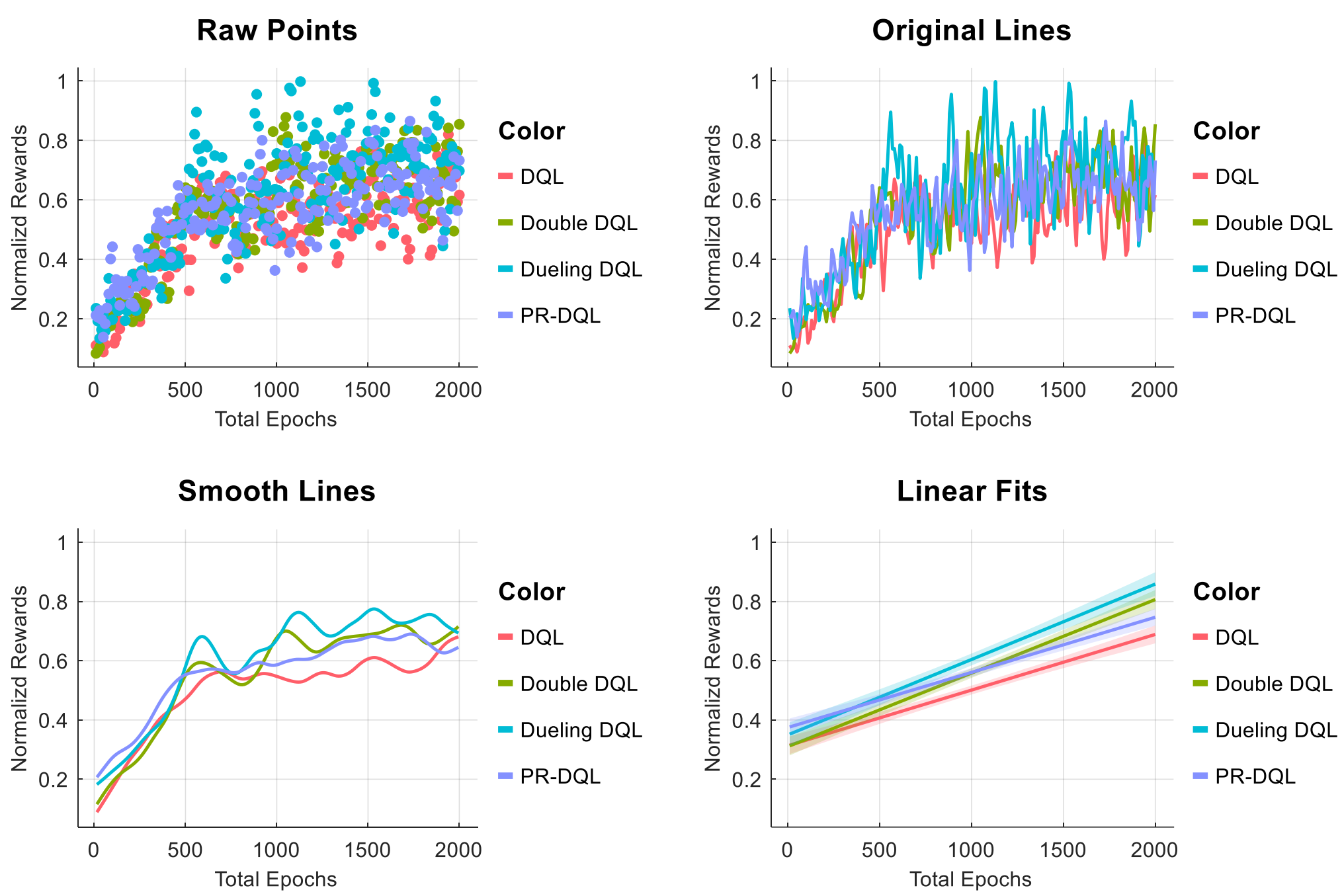

Fig. 4. Normalized reward trajectories in four DRL methods: DQL, Double DQL, Dueling DQL and PR-DQL.

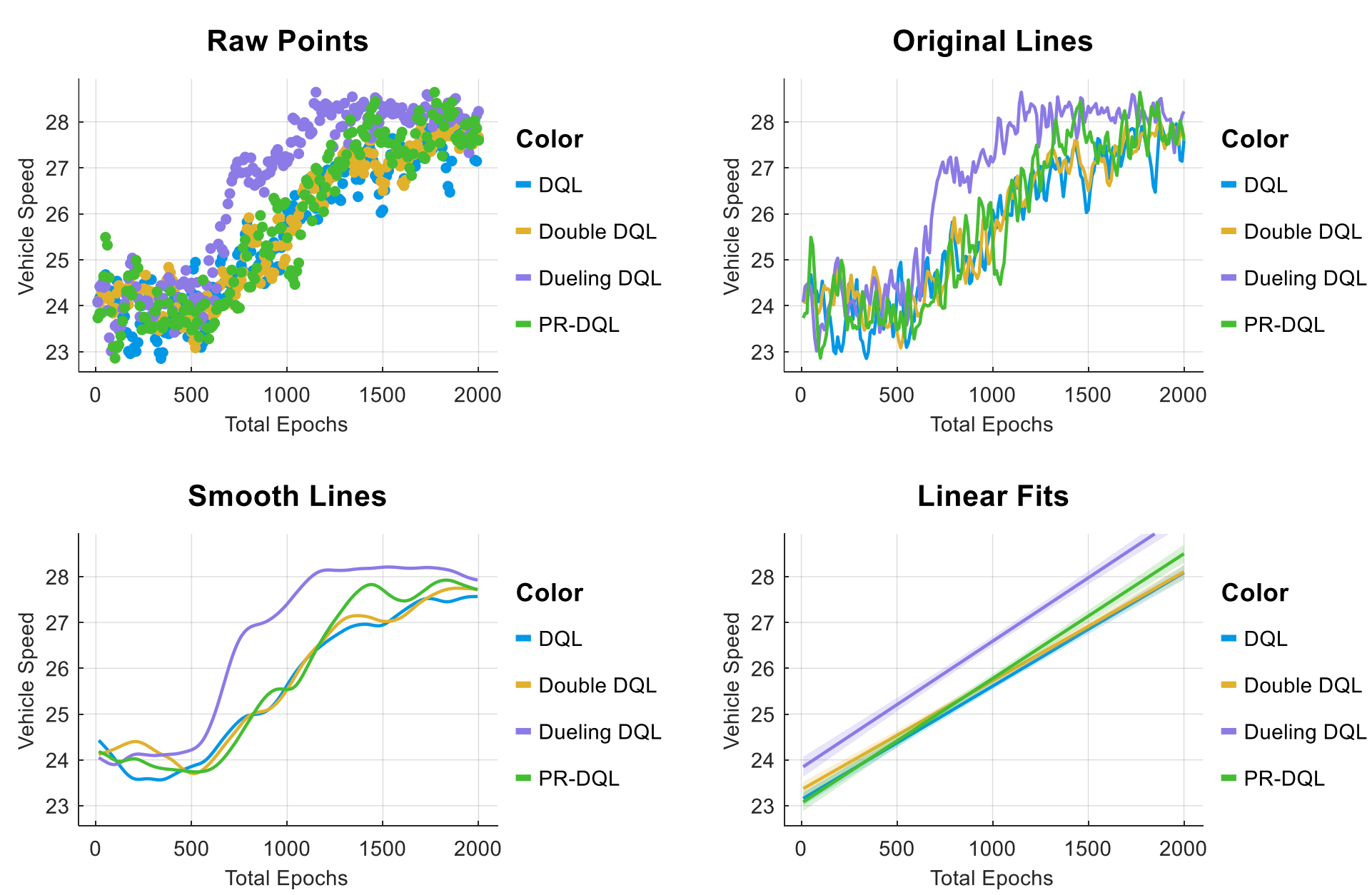

Fig. 5. Average speed curves of the ego vehicle in the four compared DQL algorithms.

In this study, the state variables chosen for analysis encompass vehicle speed and travel distance. The average speed across these four control cases is graphically depicted in Fig. 5. It's worth noting that higher vehicle speeds correspond to greater rewards, making the velocity curves a pertinent indicator of the quality of the decision-making policies. As a result of the incorporation of the advantage function within Dueling

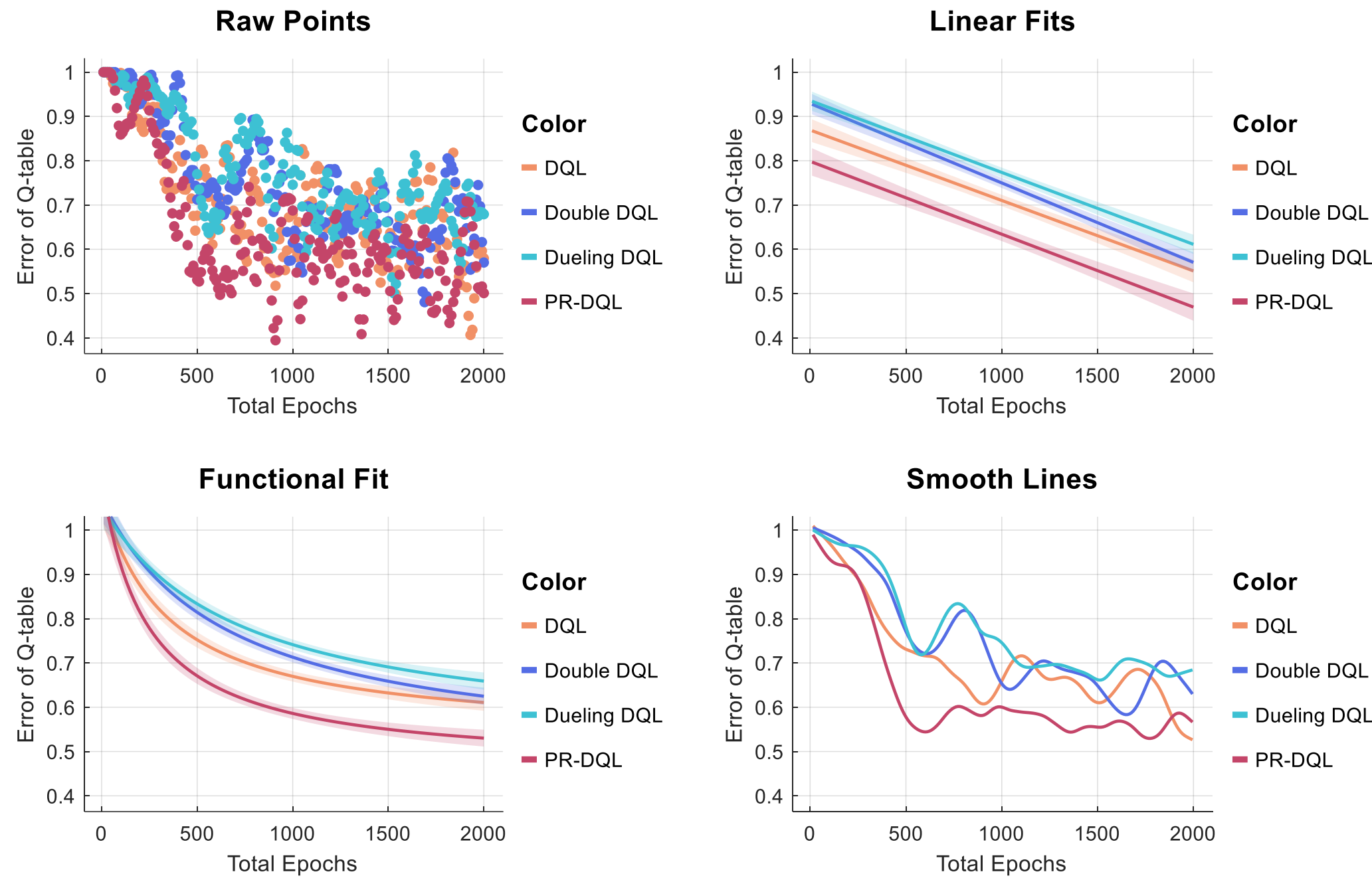

Fig. 7. Error of Q-table in four compared DRL methods to indicate the convergence rate.

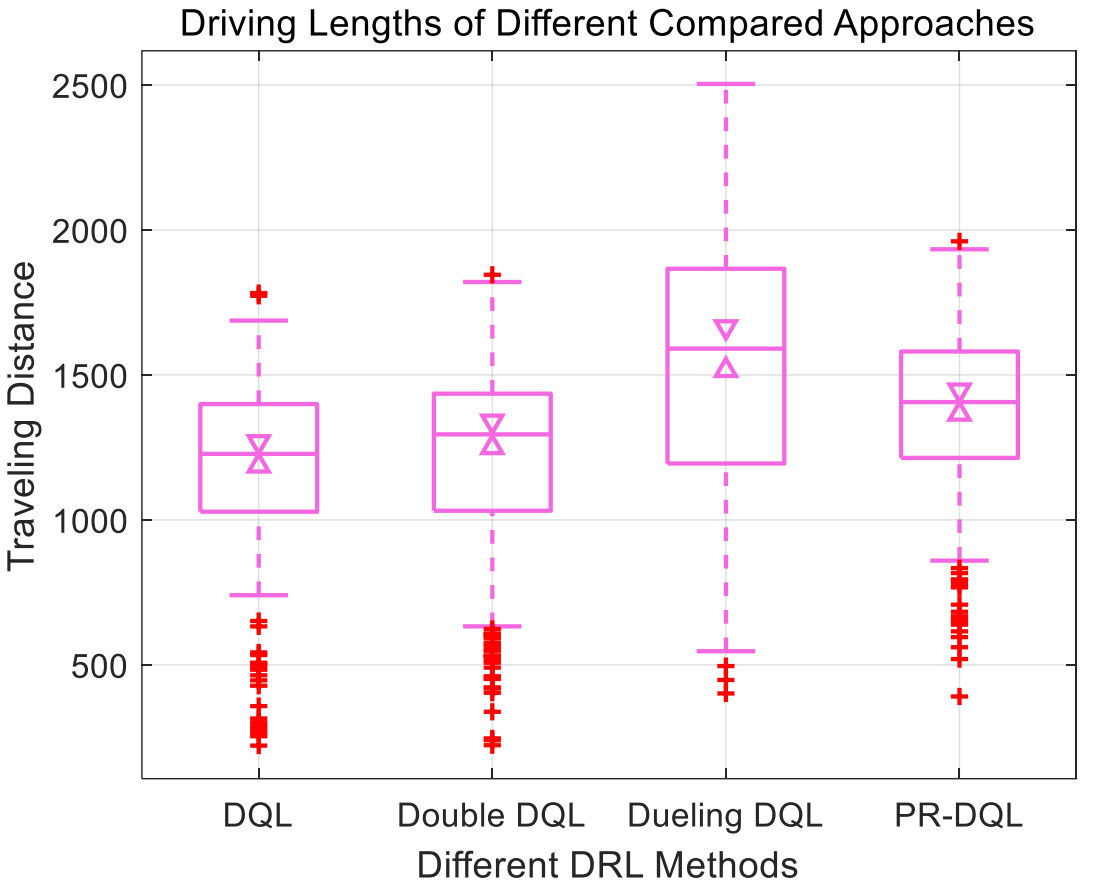

Fig. 6. Traveling distances of the ego vehicle are compared for estimation of control performance.

DQL, this algorithm consistently exhibits superior control performance. In the remaining three scenarios, PR-DQL demonstrates a more favorable speed trajectory, while Double DQL yields results similar to those of DQL.

Additionally, the travel distances covered by the ego vehicle are presented in Fig. 6. Boxplot representations are employed to encapsulate 2000 samples for each algorithm, corresponding to the number of episodes (2000). It is evident that Dueling DQL achieves the greatest travel distance, with its median value also being the highest. In this comparative analysis, PR-DQL outperforms both Double DQL and DQL. Consequently, it is apparent that the decision-making strategy employed by Dueling DQL exhibits the most efficient learning process within the context of this freeway problem.

## B. *Convergent Procedure of DRL Algorithms*

This subsection undertakes a comparison of the convergence rates exhibited by the aforementioned four Deep Reinforcement Learning (DRL) techniques. As elucidated in Section II, the fundamental distinction between Reinforcement Learning (RL) and Deep Reinforcement Learning (DRL) lies in the application of a neural network for approximating the Q-table. In the case of DQL and PR-DQL, only one neural network is utilized, whereas the other two methods involve the use of two networks. To gauge the convergence rates of different DRL algorithms, it is customary to compare the mean discrepancies within the Q-tables and cumulative rewards.

Fig. 7 illustrates the variations in Q-table errors across the four control scenarios. The primary determinant influencing these values is the training of neural networks. Notably, when examining the figures for linear fits and functional fits, it becomes apparent that PR-DQL exhibits the swiftest convergence rate. This can be attributed to the specific rules governing the selection of sample experiences within PR-DQL. In the case of DQL, which employs a single network, it outperforms both Double DQL and Dueling DQL. Conversely, the convergence process of Dueling DQL is the slowest. Consequently, although Dueling DQL demonstrates superior performance, it demands the longest training duration. It is, therefore, advisable to select different DRL methods based on the specific characteristics of the problem at hand; for instance, if time considerations are not paramount, Dueling DQL may be a suitable choice.

Furthermore, Fig. 8 presents the cumulative rewards, as defined in Equation (1), for different algorithms. These cumulative rewards represent the summation of the current reward and the discounted future rewards. Given that the reward is inherently influenced by the control actions, these curves serve as an effective means to evaluate the quality of decision-making policies. It is evident that the values achieved by Dueling DQL surpass those of the other three methods. Meanwhile, Double DQL exhibits a performance level closely resembling that of PR-DQL. The consistent upward trend observed in cumulative rewards signifies the ego vehicle's increasing familiarity with the driving environment. These findings strongly suggest that the Dueling DQL algorithm is particularly well-suited for the decision-making challenges presented by freeway scenarios

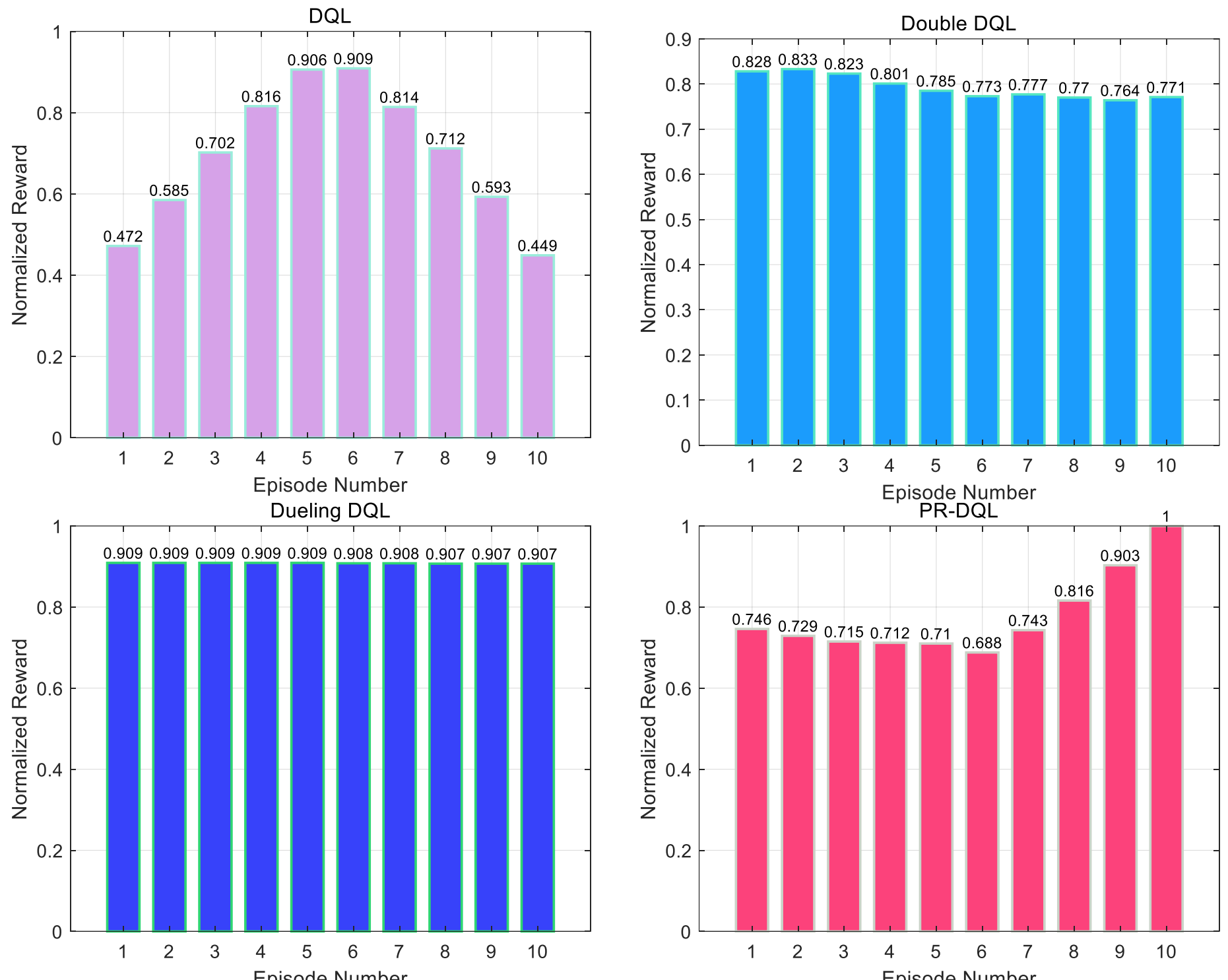

Fig. 9. Normalized values of reward in the testing experiments in four DQL algorithms.

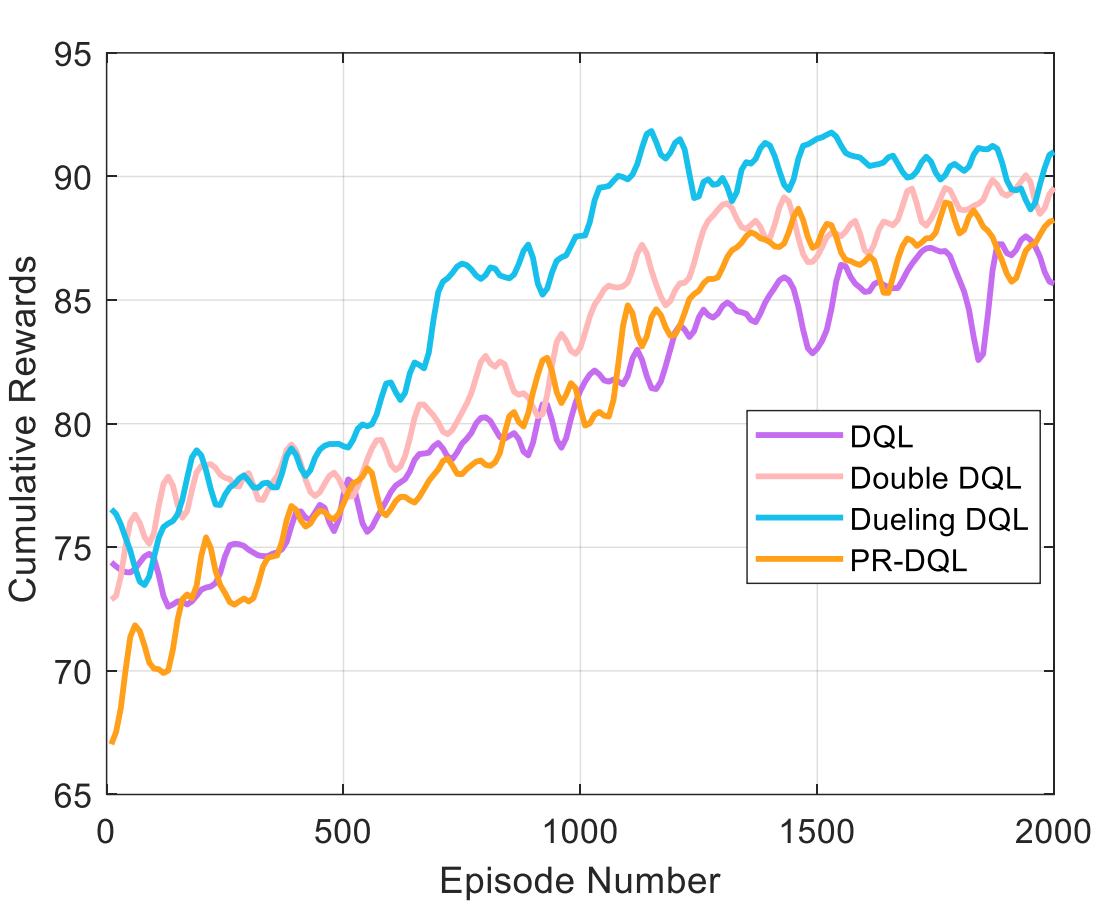

Fig. 8. Accumulated rewards in DQL, Double DQL, Dueling DQL, and PR-DQL.

### C. Testing Experiments for Trained Policies

In this subsection, we proceed to validate the performance of the four trained decision-making policies within a comparable driving scenario. The testing phase comprises a total of 10 episodes, maintaining the configuration of three lanes and 15 surrounding vehicles. However, a key distinction lies in the variability of speed and positional parameters for these vehicles, introducing an element of randomness. Consequently, the trained strategies are compelled to adapt to this dynamic driving environment. By monitoring the resultant reward values, it becomes apparent whether collisions occur or not. This analysis enables us to discern the effectiveness of these decision-making policies in addressing uncertainties within the driving context.

First and foremost, Fig. 9 provides a visual representation of the normalized rewards achieved in each testing episode. This figure yields several noteworthy observations. PR-DQL attains the highest reward value (reward=1), signifying its exemplary performance. In contrast, DQL exhibits the lowest reward value (reward=0.449), and the rewards in this instance display substantial fluctuations. Conversely, both Double DQL and Dueling DQL demonstrate remarkable stability, with Dueling DQL outperforming the former. Furthermore, the reward values within Dueling DQL remain consistently high and uniform, indicating the suitability of the DRL algorithm for addressing the decision-making challenges inherent in freeway scenarios. In the majority of cases, the three variants of DQL outperform the conventional DQL, underscoring the necessity of these adaptations.

To illustrate the distinctions among decision-making policies derived from different methods, we scrutinize the control actions chosen in four individual episodes, as presented in Fig. 10. In Fig. 9, these episodes correspond to the index values of 6, 2, 3, and 10, respectively, representing the optimal performance achieved in each method. These control actions are categorized into five indexes, denoting commands for changing lanes to the left, maintaining idle speed, changing lanes to the right, reducing speed, and increasing speed. The variation in these control actions underscores the dissimilarities in the associated decision-making policies, consequently influencing control performance and convergence rates. As established through our analysis, Dueling DQL emerges as the most effective solution for addressing the freeway decision-making problem, albeit at the expense of longer training times. In summary, the choice of the suitable DQL algorithm for a comparable decision-making problem is contingent upon the inherent complexity of the problem. For highly complex scenarios, PR-DQL is recommended, whereas for less intricate situations, Dueling DQL is the suggested choice.

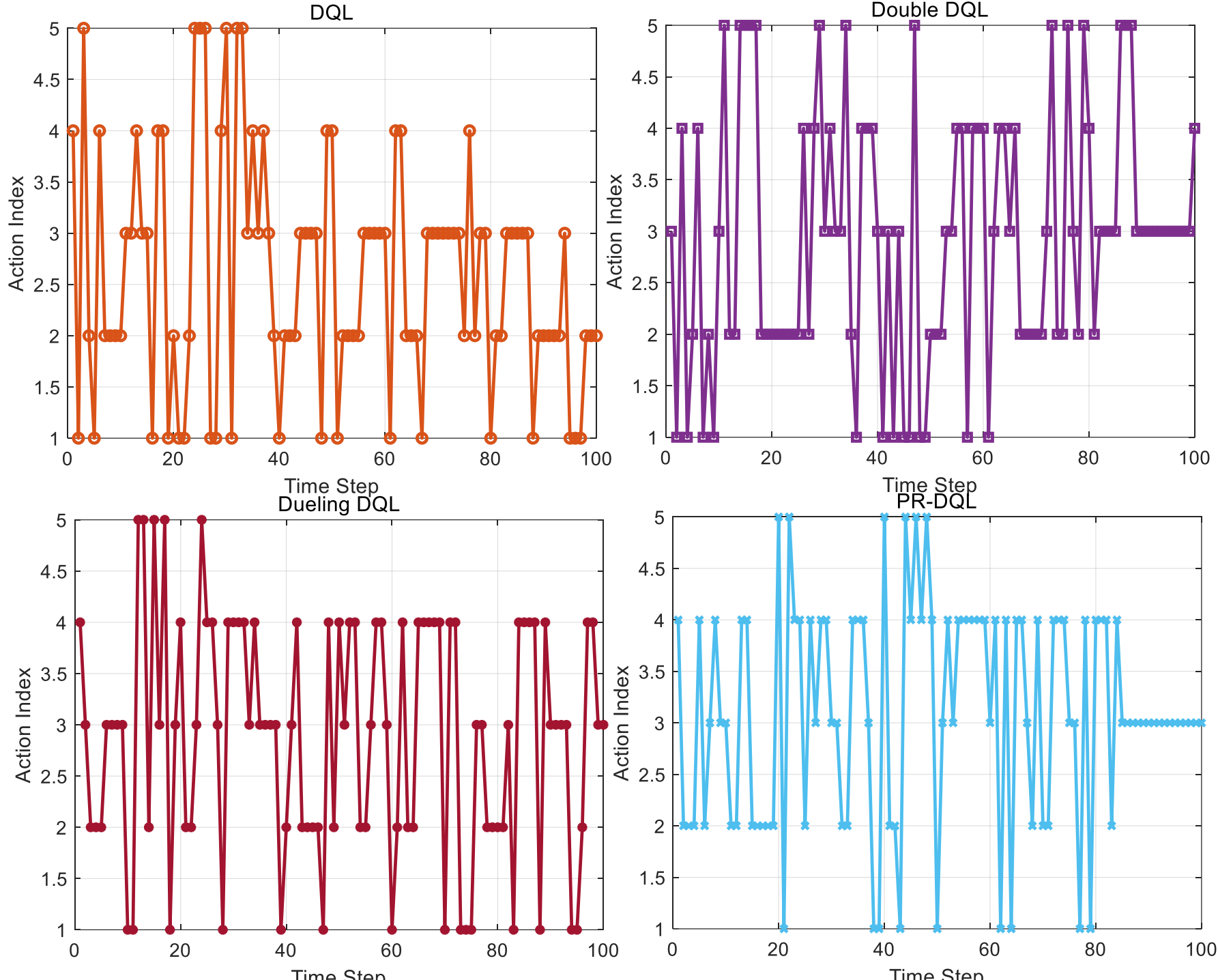

Fig. 10. Control action trajectories from four compared approaches (DQL, Double DQL, Dueling DQL and PR-DQL) in the testing environment.

## Conclusion

In this study, we have conducted a comparative analysis of four DQL algorithms in the context of decision-making for autonomous vehicles on the freeway. We initially provided an overview of the implementation workflows for these four methods. Subsequently, we established a freeway driving scenario and a bi-level control framework to govern the behavior of the ego vehicle and the surrounding vehicles. Our simulation results encompassed discussions on the learning process, convergence rates, and the outcomes of testing experiments for these four approaches. The selection of an appropriate DQL algorithm for analogous problems should be contingent upon the complexity of the research problem. In conclusion, we recommend the use of Dueling DQL and PR-DQL.

Future research directions encompass two main aspects. Firstly, the application of these decision-making policies within visualization software will be explored. Secondly, the assessment of the derived decision-making strategies will be conducted using real-world driving data.

## References

[1] A. Raj, J. A. Kumar, and P. Bansal, "A multicriteria decision making approach to study barriers to the adoption of autonomous vehicles," *Transp. Res. Pt. A-Policy Pract*, vol. 133, pp. 122-137, 2020.

[2] J. Tang, S. Liu, S. Pei, S. Zuckerman, C. Liu, and J. Gaudiot, "Teaching autonomous driving using a modular and integrated approach," *arXiv preprint arXiv:1802.09355*, 2018.

[3] T. Liu, B. Tian, Y. Ai, L. Chen, F. Liu, and D. Cao, "Dynamic States Prediction in Autonomous Vehicles: Comparison of Three Different Methods," in *Proc. 2019 IEEE ITSC*, Auckland, New Zealand, 2019, pp. 3750-3755.

[4] Q. H. Do, H. Tehrani, S. Mita, M. Egawa, K. Muto, and K. Yoneda, "Human drivers based active-passive model for automated lane change," *IEEE Intell. Transp. Syst. Mag*, vol. 9, no. 1, pp. 42-56, 2017.

[5] G. Xiong, Z. Kang, H. Li, W. Song, Y. Jin, and J. Gong, "Decision-making of Lane Change Behavior Based on RCS for Automated Vehicles in the Real Environment," in *Proc. 2018 IEEE IV*, Changshu, China, 2018, pp. 1400-1405.

[6] K. Shu, H. Yu, X. Chen, L. Chen, Q. Wang, L. Li and D. Cao, "Autonomous Driving at Intersections: A Critical-Turning-Point Approach for Left Turns," *arXiv preprint arXiv:2003.02409*, 2020.

[7] V. Sezer, "Intelligent decision making for overtaking maneuver using mixed observable Markov decision process," *J. Intell. Transport. Syst*, vol. 22, no. 3, pp. 201-217, 2017.

[8] B. Mirchevska, C. Pek, M. Werling, M. Althoff, and J. Boedecker, "High-level decision making for safe and reasonable autonomous lane changing using reinforcement learning," in *Proc. 2018 21st Int. Conf. ITSC*, Hawaii, USA, 2018, pp. 2156-2162.

[9] J. Duan, S. E. Li, Y. Guan, Q. Sun, and B. Cheng, "Hierarchical reinforcement learning for self-driving decision-making without reliance on labelled driving data," *IET Intell. Transp. Syst*, vol. 14, no. 5, pp. 297-305, 2020.

[10] F. Ye, X. Cheng, P. Wang, and C.-Y. Chan, "Automated Lane Change Strategy using Proximal Policy Optimization-based Deep Reinforcement Learning," *arXiv preprint arXiv:2002.02667*, 2020.

[11] X. Feng, J. Hu, Y. Huo, and Y. Zhang, "Autonomous Lane Change Decision Making Using Different Deep Reinforcement Learning Methods," in *Proc. 19th. CICTP*, Nanjing, China, 2019, pp. 5563-5575.

[12] T. Liu, B. Huang, Z. Deng, H. Wang, X. Tang, X. Wang and D. Cao, "Heuristics-oriented overtaking decision making for autonomous vehicles using reinforcement learning", *IET Electr. Syst. Transp*, 2020.

[13] R. Sutton, and A. Barto, in *Reinforcement learning: An introduction*, 2th ed. Cambridge, Massachusetts, USA:MIT press, 2018.

[14] T. Liu, X. Hu, W. Hu, Y. Zou, "A heuristic planning reinforcement learning-based energy management for power-split plug-in hybrid elec-

tric vehicles," *IEEE Trans. Ind. Inform*, vol. 15, no. 12, pp. 6436-6445, 2019.
[15] A. Alizadeh, M. Moghadam, Y. Bicer, N. Ure, U. Yavas, and C. Kurtulus, "Automated Lane Change Decision Making using Deep Reinforcement Learning in Dynamic and Uncertain Highway Environment," in *Proc. 2019 IEEE ITSC*, Auckland, New Zealand, 2019, pp. 1399-1404.
[16] X. Hu, T. Liu, X. Qi, and M. Barth, "Reinforcement learning for hybrid and plug-in hybrid electric vehicle energy management: Recent advances and prospects," *IEEE Ind. Electron. Mag*, vol. 13, no. 3, pp. 16-25, 2019.
[17] T. Liu, X. Tang, H. Wang, H. Yu, and X Hu, "Adaptive Hierarchical Energy Management Design for a Plug-in Hybrid Electric Vehicle," *IEEE Trans. Veh. Technol*, vol. 68, no. 12, pp. 11513-11522, 2019.
[18] T. Liu, B. Wang, and C. Yang, "Online Markov Chain-based energy management for a hybrid tracked vehicle with speedy Q-learning," *Energy*, vol. 160, pp. 544-555, 2018.
[19] T. De Bruin, in *Sample Efficient Deep Reinforcement Learning for Control*, 2020.
[20] V. Mnih, K. Kavukcuoglu, D. Silver, A. Graves, I. Antonoglou, D. Wierstra and M. Riedmiller, "Playing atari with deep reinforcement learning," *arXiv preprint arXiv:1312.5602*, 2013.
[21] H. Van Hasselt, A. Guez, and D. Silver, "Deep reinforcement learning with double q-learning," in *Proc. 30th AAAI-16*, Arizona, USA, 2016, pp. 2094-2100.
[22] Z. Wang, T. Schaul, M. Hessel, H. Hasselt, M. Lanctot, and N. Freitas, "Dueling network architectures for deep reinforcement learning," in *Proc. Int. conf. machine learning*, New York, USA, 2016, pp. 1995-2003.
[23] T. Schaul, J. Quan, I. Antonoglou, and D. Silver, "Prioritized experience replay," *arXiv preprint arXiv:1511.05952*, 2015.
[24] C. Hoel, K. Driggs-Campbell, K. Wolff, L. Laine, and M. J. Kochenderfer, "Combining planning and deep reinforcement learning in tactical decision making for autonomous driving," *IEEE Trans. Intell. Veh*, vol. 5, no. 2, pp. 294-305, 2019.
[25] M. Treiber, A. Hennecke, and D. Helbing, "Congested traffic states in empirical observations and microscopic simulations," *Phys. Rev. E*, vol. 62, no. 2, p. 1805, 2000.
[26] A. Kesting, M. Treiber, and D. Helbing, "General lane-changing model MOBIL for car-following models," *Transp. Res. Record*, vol. 1999, no. 1, pp. 86-94, 2007.
[27] P. Polack, F. Altché, B. d'Andréa-Novel, and A. de La Fortelle, "The kinematic bicycle model: A consistent model for planning feasible trajectories for autonomous vehicles," in *Proc. 2017 IEEE IV,* California, USA, 2017, pp. 812-818.
[28] L. Edouard, "An environment for autonomous driving decision-making," https://github.com/ eleurent/highway-env, *GitHub*, 2018.

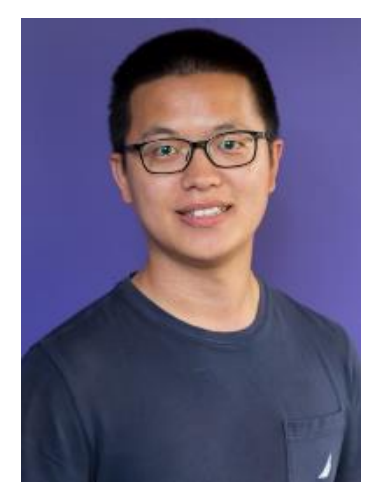

**Teng Liu** (M'2018) received the B.S. degree in mathematics from Beijing Institute of Technology, Beijing, China, 2011. He received his Ph.D. degree in automotive engineering from Beijing Institute of Technology (BIT), Beijing, in 2017. His Ph.D. dissertation, under the supervision of Pro. Fengchun Sun, was entitled "Reinforcement learning-based energy management for hybrid electric vehicles." He worked as a research fellow in Vehicle Intelligence Pioneers Ltd from 2017 to 2018. Dr. Liu worked as a postdoctoral fellow at the Department of Mechanical and Mechatronics Engineering, University of Waterloo, Ontario N2L3G1, Canada from 2018 to 2020. Now, he is a member of IEEE VTS, IEEE ITS, IEEE IES, IEEE TEC, and IEEE/CAA.
Dr. Liu is now a Professor at the Department of Automotive Engineering, Chongqing University, Chongqing 400044, China. He has more than 8 years' research and working experience in renewable vehicle and connected autonomous vehicle. His current research focuses on reinforcement learning (RL)-based energy management in hybrid electric vehicles, RL-based decision making for autonomous vehicles, and CPSS-based parallel driving. He has published over 40 SCI papers and 15 conference papers in these areas. He received the Merit Student of Beijing in 2011, the Teli Xu Scholarship (Highest Honor) of Beijing Institute of Technology in 2015, "Top 10" in 2018 IEEE VTS Motor Vehicle Challenge and sole outstanding winner in 2018 ABB Intelligent Technology Competition. Dr. Liu is a workshop co-chair in 2018 IEEE Intelligent Vehicles Symposium (IV 2018) and has been reviewers in multiple SCI journals, selectively including IEEE Trans. Industrial Electronics, IEEE Trans. on Intelligent Vehicles, IEEE Trans. Intelligent Transportation Systems, IEEE Transactions on Systems, Man, and Cybernetics: Systems, IEEE Transactions on Industrial Informatics, Advances in Mechanical Engineering.

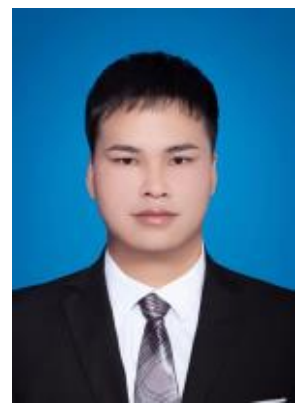

**Yuyou Yang** is currently pursuing the M.Sc. degree in the College of Mechanical and Vehicle Engineering, Chongqing University, Chongqing, China. He received the B.E. degree in College of Mechanical Engineering from the Chongqing University in 2021. His research interests focus on path planning and decision-making of autonomous vehicles.

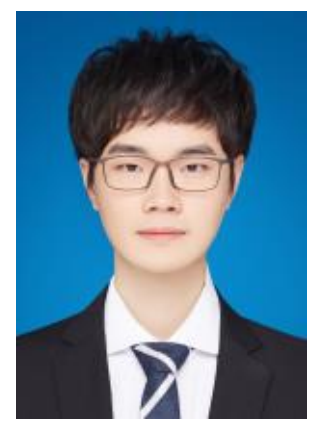

**Wenxuan Xiao** is currently pursuing the M.Sc. degree in the College of Mechanical and Vehicle Engineering, Chongqing University, Chongqing, China. He received the B.E. degree in College of Automotive Engineering of Chongqing University in 2021. His research interests focus on path planning and decision control for autonomous driving in highway ramp and intersection scenarios.

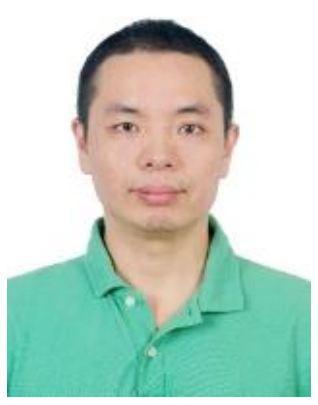

**Xiaolin Tang** received a B.S. in mechanics engineering and an M.S. in vehicle engineering from Chongqing University, China, in 2006 and 2009, respectively. He received a Ph.D. in mechanical engineering from Shanghai Jiao Tong University, China, in 2015. From August 2017 to August 2018, he was a visiting professor of Department of Mechanical and Mechatronics Engineering, University of Waterloo, Waterloo, ON, Canada. He is currently an Associate Professor at the Department of Automotive Engineering, Chongqing University, Chongqing, China. He is also a committeeman of Technical Committee on Vehicle Control and Intelligence of Chinese Association of Automation (CAA). He has led and has been involved in more than 10 research projects, such as National Natural Science Foundation of China, and has published more than 30 papers. His research focuses on Hybrid Electric Vehicles (HEVs), vehicle dynamics, noise and vibration, and transmission control.

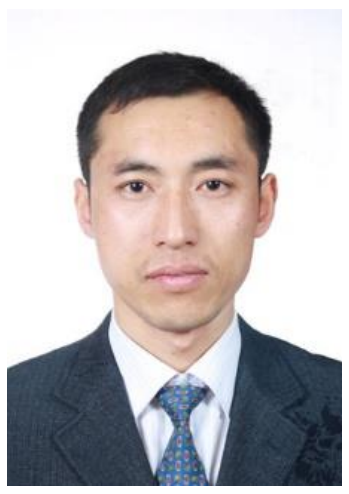

Dr. Mingzhu Yin is the Professor and Chair of Cancer Early Diagnosis Center (CEDC), Clinical Research Center (CRC), Clinical Pathology Center (CPC), Chongqing University Three Gorges Hospital, Chongqing University, Director of Translational Medicine Research Center (TMRC), Chongqing University, and Adjunct Assistant Professor of Department of Pathology, Yale University School of Medicine. Dr. Yin completed his first postdoctoral training on genetics at the National Institute on Aging, National Institutes of Health in 2011, and the second postdoctoral training on pathology at Yale University in 2013. He received his Ph.D. degree of Integration traditional Chinese and Western Medicine from China Pharmaceutical University in 2018. After returning to China in 2018, he established the largest domestic single-cell sequencing and analysis technology platform and database. He was approved as The Top Young Talents of Ten Thousand Talents program of Organization Department of the Central Committee of the of the CPC in 2019, Hunan Outstanding Youth Fund in 2021, and the Leading talent of the Ministry of Industry and Information Technology in 2022.